\documentclass[letterpaper, 10 pt, conference, final]{ieeeconf}  

\IEEEoverridecommandlockouts                              

\usepackage{amsmath, amssymb}
\usepackage{gensymb}
\usepackage{algorithm}
\usepackage{algorithmicx}
\usepackage{algpseudocode}
\usepackage{booktabs}
\usepackage{multirow}
\usepackage{graphicx}
\usepackage{wrapfig}
\usepackage{float}
\usepackage{subcaption}
\usepackage{comment}
\usepackage{xcolor}

\newcommand{\code}[1]{\texttt{#1}}

\title{\LARGE \bf
Graph-Structured Policy Learning for Multi-Goal Manipulation Tasks
}

\author{David Klee$^{1,*}$, Ondrej Biza$^{1}$ and Robert Platt$^{1}$
\thanks{$^{1}$Khoury College of Computer Sciences, Northeastern University, 
        Boston, MA 02215 USA}
\thanks{$^{*}$Corresponding author {\tt\small klee.d@northeastern.edu}}%
}

\begin{document}

\maketitle
\thispagestyle{empty}
\pagestyle{empty}

\begin{abstract}

Multi-goal policy learning for robotic manipulation is challenging.  Prior successes have used state-based representations of the objects or provided demonstration data to facilitate learning. In this paper, by hand-coding a high-level discrete representation of the domain, we show that policies to reach dozens of goals can be learned with a single network using Q-learning from pixels.  The agent focuses learning on simpler, local policies which are sequenced together by planning in the abstract space.  We compare our method against standard multi-goal RL baselines, as well as other methods that leverage the discrete representation, on a challenging block construction domain. We find that our method can build more than a hundred different block structures, and demonstrate forward transfer to structures with novel objects.  Lastly, we deploy the policy learned in simulation on a real robot.

\end{abstract}

\section{INTRODUCTION}

\begin{figure}[t]
  \centering
  \includegraphics[width=\linewidth]{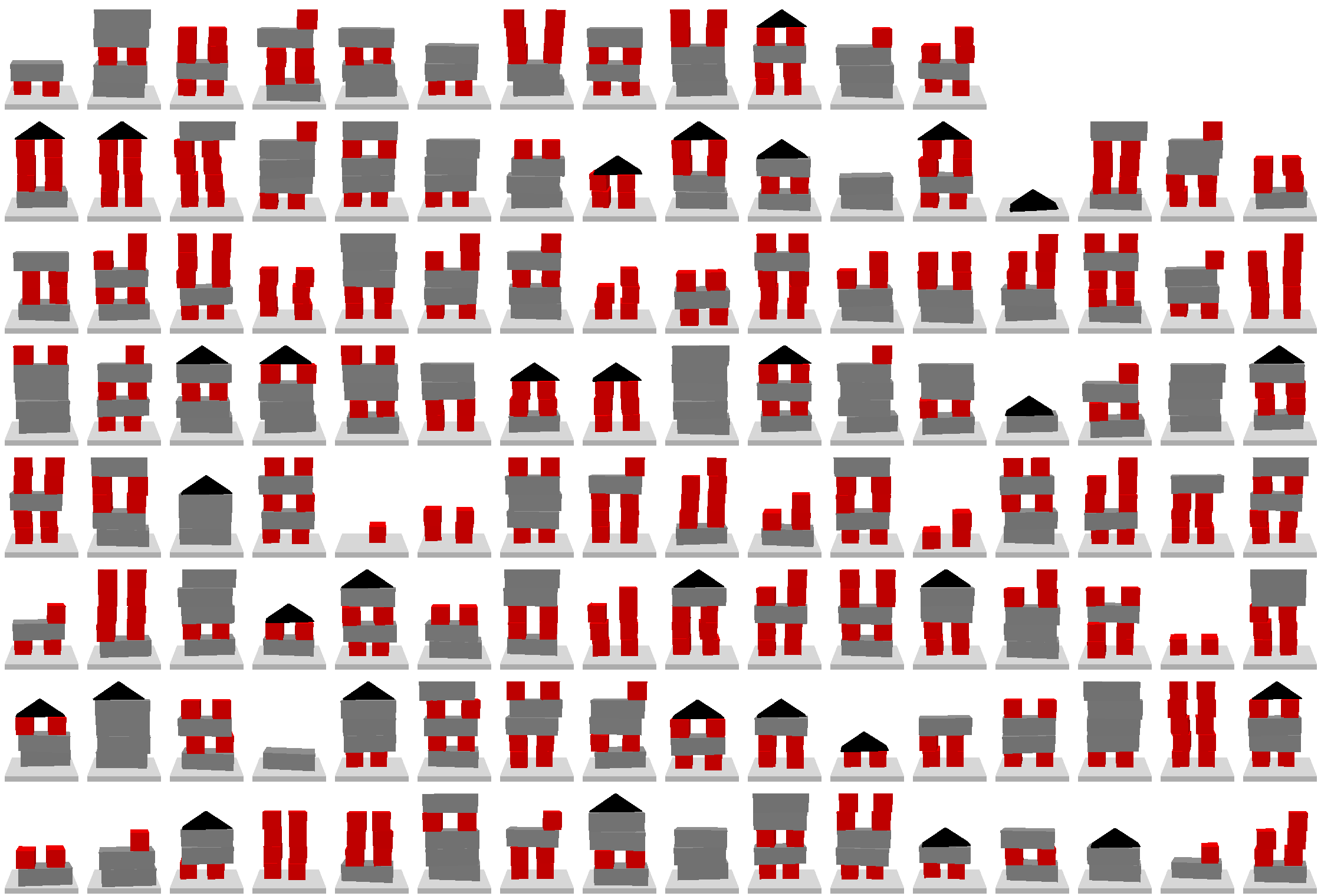}
  \caption{Our multi-goal framework enables us to build any structure shown above by sequencing learned subgoal policies. All structures shown were built by a single agent trained with our method in the simulator.}
  \label{fig:height_5_structures}
  \vspace{-1em}
\end{figure}

Multi-goal policy learning is important to robotic manipulation because it offers the possibility that an agent could learn to solve a large number of different manipulation problems in a single domain. For example, in a block construction domain, we would like to learn a set of policies that could build any structure shown in Fig.~\ref{fig:height_5_structures}. 
Unfortunately, standard methods for multi-goal policy learning have not yet scaled well to problem domains like that shown here \cite{plappert2018multi}. Standard goal conditioned policy learning quickly runs into network capacity limits -- it is difficult for a single neural network to encode policies for hundreds of different construction tasks. Moreover, standard methods ignore a key characteristic here -- that each of the construction problems shown in Fig.~\ref{fig:height_5_structures} shares common substructure with a subset of other tasks within the group. 
This is a situation where planning methods have the potential to work well. If we can accurately segment the objects in the world and estimate their position and geometry, then a variety of planning methods exist that can be used to find goal-reaching solutions. Planning methods have an advantage here because they have the ability to solve a large number of tasks in a given domain. Unfortunately, since these methods do not learn end-to-end, they have no ability to improve based on experience.

In this paper, we attempt to strike a balance between multi-goal policy learning and planning with a simple two-level hierarchy that solves discrete planning problems at a high level and executes learned policies at a low level. The high level operates like a standard discrete planner with a hand-coded discrete state space and transition function, not unlike the task planner in standard TAMP methods~\cite{kaelbling2010hierarchical,srivastava2014combined}. Transitions at the high level correspond to goal-conditioned policy learning problems at the low level. Our algorithm finds a path in the high-level discrete state space and then attempts to learn policies (or execute already-learned policies) that traverse the path. Whereas standard multi-goal policy learning methods attempt to learn end-to-end policies for all goals of interest, our method learns a large number of subgoal-reaching policies that are each valid only in a small region of state space (similar to TDM~\cite{pong2018temporal}). This helps us learn policies for a large number of subgoals without exceeding network capacity limits. Note that this is not a hierarchical RL method because we are hand-coding the hierarchy instead of learning it. However, in many robotic manipulation problems, we believe it is relatively easy to hand-code this kind of structure compared to solving the challenging perception problems that would otherwise be needed in standard planning. 

Our main contribution is to show that by combining a hand-coded, structured goal space with multi-goal policy learning, it is possible to solve an otherwise extremely challenging robotic manipulation problem.  We propose a structured policy representation for manipulation that can represent policies over a large goal space compactly and can generate an implicit curriculum.  To our knowledge, no other multi-goal method can learn a policy that can solve the array of 124 block construction problems shown in Fig.~\ref{fig:height_5_structures} using image-based state without demonstration data.  Additionally, we show that the structured goal space can be expanded after training to perform forward transfer to block structures with novel objects.

\section{RELATED WORK}

\noindent\underline{Goal-conditioned Reinforcement Learning:} An RL agent is conditioned on a goal, which can be a particular state (e.g. move a block to an $(x, y)$ position) or a more abstract notion of a goal (e.g. build a tower from four blocks). Methods in this area often adaptively choose goals to explore in order to speed up learning. In the context of model-free RL, Andrychowicz et al. \cite{andrychowicz2017hindsight} consider every state a goal, and train their agent on achieved goals via relabelling. \cite{zhao18energy,zhao19entropy} extended the framework to increase the diversity of explored goals and replayed experiences.  In contrast, this work addresses problems where the goals are a limited and difficult to achieve area of the state space, so relabelling does not generate a useful reward signal.
Several works \cite{li2020towards, funk2022learn2assemble, lin2022efficient} use object-factored models to build large block structures, whereas our approach learns from image observations. Other works \cite{nair18visual,nasiriany2019planning, janner2018reasoning} have performed model-based planning in the learned latent space to solve the block construction task from image observations.

\noindent
\underline{Automatic curricula for RL agents:} Previous works have explored automatic curriculum generation, usually for the purpose of breaking down a difficult task into easier subtasks. Prior work has used a generative model \cite{florensa18autmatic} or ensemble of value functions \cite{zhang2020automatic} to generate goals that an agent can more readily learn. \cite{florensa17reverse} manipulated the starting state of an agent so that it is likely to reach the goal with its current abilities. Other works considered a framing with a teacher and a student agent, where a teacher either learns to select tasks for the student, or the student attempts to imitate the teacher \cite{sukhbaatar18intrinsic,matiisen20teacher,mukherjee2021reactive}.

\noindent
\underline{Learning and planning with symbols:} Similarly to our structured (symbolic) goal space, Zhang et al. \cite{zhang18composable} considered a setting where each state has a ground-truth symbolic description. Their symbols do not pertain to goals specifically, but to other characteristics of states, such as two blocks being next to each other in a grid-world. The goal of their work is to learn low-level policies and a transition function for a high-level planner, whereas we are specifically interested in solving a large number of goals.  Symbolic task descriptions have also been considered in the context of learning of composable skills \cite{kaelbling17learning,james20learning}.

\noindent
\underline{Options framework}: Learning low-level policies to perform temporally extended tasks has been explored using the options framework~\cite{stolle2002learning}.  Option policies can be learned end-to-end ~\cite{bacon2017option,vezhnevets2017feudal}, but the state-of-the-art is currently limited to a low number of options.  In the context of the proposed method, the high-level planner takes the role of the initiation set, deciding which policy to deploy, and termination occurs when the subgoal is reached. Konidaris et al.~\cite{konidaris2018skills} learned options defined for a set of abstract subgoals, and performed planning over options for a mobile manipulation task.

\section{Approach}

\subsection{Manipulation as a Goal-Conditioned MDP in a Spatial Action Space}
\label{sect:mdp}

We formulate the manipulation problem as a deterministic goal-conditioned Markov Decision Process (MDP) in a spatial action space. Typically, when formulating a robotics problem as an MDP, the action space is over end effector or joint velocities~\cite{watter15embed,levine18learning}. But, in a spatial action space, the action space includes a subset of $SE(2)$ (or $SE(3)$) where the spatial variables denote the target pose for an arm motion (rather than a target velocity). The spatial action space gives our agent access to temporally abstract actions (i.e. end-to-end reaching actions), simplifying the learning problem.

\noindent
\underline{Goal Conditioned MDP:} A goal-conditioned Markov decision process $\mathcal{M} = (S,A,G,R_g,T,\rho,\gamma)$ has a state space $S$, an action space $A$, a family of goal sets $G \subset 2^S$, an unknown transition function $T : S \times A \rightarrow Pr(S)$, a goal conditioned reward function $R_g : S \rightarrow \{-1,0\}, \; g \in G$, a distribution of initial states $\rho$, and the discount factor $\gamma$. In this work, the reward function is sparse with $R_g(s,a) = -1$ if $s \notin g$ and $R_g(s,a) = 0$ otherwise. 

\noindent
\underline{Visual State Space:} The state space $S = S_{world} \times S_{robot}$ is the cross product of the state of the world and the state of the robot (robot state includes objects held by the robot). The state of the world is expressed as a single $n$-channel image $I \in S_{world} = \mathbb{R}^{n \times h \times w}$ that contains all relevant objects in a scene. The state of the robot could be arbitrary, but in this paper it is encoded by the ``in-hand'' image $H \in S_{robot} = \mathbb{R}^{n \times h' \times w'}$ ($h' < h$ and $w' < w$) that shows objects currently grasped by the robot, and a boolean flag $b$ indicating if the robot is holding an object (i.e. was gripper able to fully close on last \textsc{pick}). If the robot is not holding anything, then the pixels in $H$ are zero. $H$ is generated by storing an $h' \times w'$ image crop from $I$ just prior to executing the last pick action. An example of $I$ and $H$ is shown in Fig.~\ref{fig:model_architecture}.

\noindent
\underline{Spatial Action Space:} The action space $A = A_{sp} \times A_{arb}$ is the cross product of the spatial component of action $a_{sp} \in A_{sp}$ and additional arbitrary action variables $a_{arb} \in A_{arb}$. Component $a_{sp}$ encodes the position to which the robot is to move its hand while $a_{arb}$ encodes information about how the robot is to move or what will happen after the move is complete. In this paper, $A_{sp}$ is the space of $x,y$ positions where the hand is moved to and $A_{arb} = \{\textsc{pick},\textsc{place}\}$ denotes whether this is a pick action or a place action. If $a_{arb} = \textsc{pick}$, then the robot closes the gripper after moving to the position $s_{sp}$. If $a_{arb} = \textsc{place}$, then the robot opens its fingers after completing the move. 

\noindent
\underline{Learning Objective:} Given the goal-conditioned MDP formulated above, our objective is to find a goal-conditioned policy that achieves optimal expected returns with respect to a probability distribution over goals, $\Lambda : G \mapsto \mathbb{R}$. $\Lambda$ can be arbitrary, for example allocating all probability mass to a single goal or evenly over a subset of goals.

\subsection{Abstract Structure}
\label{sect:abstract-structures}

We structure learning using a hand-coded discrete graph representation of the problem that we call the \textit{goal graph}. First, our agent solves a discrete planning problem over the goal graph. Then, it executes a sequence of goal-conditioned policies that move the agent along a planned path. 

\noindent
\underline{Goal Graph:} The goal graph is a graph $\Gamma = (\bar{S},\bar{T})$, where $\bar{S}$ is a set of abstract states and $\bar{T}$ is a set of edges. Each underlying state $s \in S$ maps onto an abstract state $\bar{s} \in \bar{S}$ via the abstraction function $f : S \rightarrow \bar{S}$, which we assume is known to the agent, i.e. $f$ is provided by the system designer. The graph must be such that each goal set $g \in G$ in the goal-conditioned MDP $\mathcal{M}$ corresponds to the preimage of an abstract state. That is: $\forall g \in G$, $\exists \bar{s} \in \bar{S}$ such that $f^{-1}(\bar{s}) = g$ (where $f^{-1}(\bar{s}) = \{s \in S | f(s) = \bar{s} \}$ denotes the preimage of $\bar{s}$ under $f$). The connectivity of $\Gamma$ encodes how our agent will attempt to reach end goals and should reflect the connectivity of the underlying problem. If an edge $(\bar{s}_1, \bar{s}_2) \in \bar{T}$ exists in the graph, then our agent may attempt to learn a policy that traverses that edge. We assume that each edge in the graph can be traversed in at most $k$ time steps.
Fig.~\ref{fig:block_graph} illustrates a portion of the goal graph for the block construction domain that we explore in this paper. Each vertex in the graph is an abstract state $\bar{s}$ which corresponds to a qualitative block structure. This abstract state maps to a set of underlying states in the MDP via the preimage of $f$, $f^{-1}(\bar{s}) \subset S$.

\subsection{Plan Execution}

Control happens at two levels in our framework. 

\begin{wrapfigure}[15]{r}{0.2\textwidth}
  \vspace{-0.25cm}
  \begin{center}
  \includegraphics[width=0.16\textwidth]{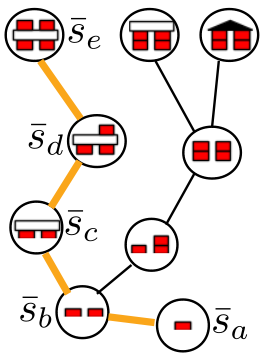}
  \end{center}
  \caption{Subgraph of the full goal graph for the block construction.}
\label{fig:block_graph}
\end{wrapfigure}

\noindent
\underline{Planning at the high level:} We use graph search to find a path through the goal graph that reaches the end-goal abstract state. This is illustrated in Fig.~\ref{fig:block_graph} by the orange line from $\bar{s}_a$ to $\bar{s}_e$. Each successive vertex on the path denotes an abstract state (qualitative block structure) that can be reached from the last in at most $k$ time steps ($k=2$ for the block construction domain).

\noindent
\underline{Policy execution at the low level:} We learn and execute goal conditioned policies using deep Q-learning \cite{mnih2015human}. We learn a $Q$ function $Q(s,a,\bar{s})$, where $\bar{s}$ is the abstract subgoal state. Let $\pi_{Q}(\bar{s})$ denote the policy induced by $Q$ with $\bar{s}$ as a goal -- the \textit{subgoal policy} for $\bar{s}$. Subgoal policies are executed for the sequence of abstract subgoal states on the path found using graph search. In Fig.~\ref{fig:block_graph} for example, the agent is initialized in a state $s \in f^{-1}(\bar{s}_a)$ and parameterized with $\bar{s}_e$ as a goal. The plan generated at the high level is a sequence of abstract states $\bar{s}_b, \bar{s}_c, \bar{s}_d, \bar{s}_e$. Starting from $s \in f^{-1}(\bar{s}_a)$, the agent executes $\pi_{Q}(\bar{s}_b)$. If the agent reaches a state $s \in f^{-1}(\bar{s}_b)$, $\pi_{Q}(\bar{s}_b)$ terminates and the agent begins executing $\pi_{Q}(\bar{s}_c)$. This continues until reaching a state $s \in f^{-1}(\bar{s}_e)$.

\noindent
\underline{Algorithm:} Algorithm~\ref{alg:alg1} summarizes the process described above. At the beginning of each episode, Step 4 samples an abstract end-goal $\bar{s}_{end}$ from $\Lambda$ ($\bar{s}_e$ in Fig.~\ref{fig:block_graph}). Step 7 does discrete planning (orange line in Fig.~\ref{fig:block_graph}) and returns the next abstract subgoal state on the path to $\bar{s}_{end}$.
Step 8 executes the low-level policy for this abstract subgoal. If the agent fails to reach $\bar{s}_{next}$ within a maximum number of time steps, then the episode terminates in failure. 

\begin{figure}
    \begin{minipage}{0.48\textwidth}
        \begin{algorithm}[H]
        \caption{Execute Plan}
        \label{alg:alg1} 
        \begin{algorithmic}[1]
            \Require  $\code{env}$ : environment, $\Gamma$ : goal graph, $\Lambda$ : distribution over end goals, $f$ : abstraction function.
            \State $Q_\theta.\code{initialize()}$
    \For{\textbf{each} $episode$}
            \State $\code{env.reset}()$
            \State $\bar{s}_{end} \sim \Lambda$
            \While{$f(s) \neq \bar{s}_{end}$}
                \State $s \Leftarrow \code{env}.\code{get\_state}()$ 
                \State $\bar{s}_{next}$ $\Leftarrow$ $\Gamma.\code{get\_next\_subgoal}(f(s),\bar{s}_{end})$
                \State $s$ $\Leftarrow$  $Q_\theta.\code{execute\_policy}(\code{env},\bar{s}_{next})$
                \If{$f(s) \neq \bar{s}_{next}$}
                    \State break
                \EndIf
                \EndWhile
        \EndFor
        \end{algorithmic}
      \end{algorithm}
    \end{minipage}
\end{figure}

\subsection{Policy Learning} 
\label{sect:policy_learning}

Subgoal policies are learned on-line during execution of Algorithm~\ref{alg:alg1} inside of $Q_\theta.\code{execute\_policy}(\code{env},\bar{s}_{next})$ in Line 8. This is standard Double DQN \cite{van2016deep} with $\gamma=1$, and a hard target update performed every 1000 optimization steps. The subgoal policy executes for at most $m$ time steps ($m=4$ in our experiments) and terminates early as soon as the subgoal state is reached. During training, the agent follows an $\epsilon$-greedy exploration strategy with a linear annealing schedule. The value of $\epsilon$ controls both the probability of a random action and the amount of random Gaussian noise added to the output of the Q-network.

\noindent
\underline{Sequenced subgoal policy learning:} Notice that since Algorithm~\ref{alg:alg1} does not do any goal relabeling, it does not learn all subgoal policies at the same time. Instead, it learns subgoal policies in the same order as they appear in plans generated in Step 7. In the example shown in Fig.~\ref{fig:block_graph} where the graph planner generates the subgoal sequence $\bar{s}_a, \bar{s}_b, \bar{s}_c, \bar{s}_d, \bar{s}_e$, our agent does not begin learning the policy for subgoal $\bar{s}_c$ until the $\bar{s}_b$ policy is learned sufficiently well to enable the agent to reach $\bar{s}_b$. If the agent never reaches $\bar{s}_b$, then the subgoal policy for $\bar{s}_c$ never executes and $\pi_{Q}(\bar{s}_c)$ is never trained. This approach has a couple of important effects. First, as we describe below, the policies that are learned for early subgoals condition the distribution of states over which we learn later subgoals. Second, the approach does not waste any time learning policies that are not relevant to any task, i.e. it does not learn policies for subgoals that are not on the planned path for any end-goals that have positive probability under $\Lambda$. 

\noindent
\underline{Learning subgoal policies over a limited radius:} Perhaps the most important outcome of sequencing subgoal execution as described in Algorithm~\ref{alg:alg1} is that learning is focused on states from which the subgoal can be reached in a small number of steps. This constraint is not explicitly enforced. However, since each subgoal in the goal graph $\Gamma$ can be reached in at most $k$ steps from adjacent subgoals, most learning will occur in the neighborhood of this $k$ step between the two subgoals. This is important because it reduces the network capacity that the $Q$ function would otherwise need. For example, we can learn subgoals policies for most of the 124 subgoal states shown in Fig.~\ref{fig:height_5_structures} without exceeding the capacity of a standard $U$-net model architecture. 

\noindent
\underline{Learning the abstraction function $f$:} In Section \ref{sect:mdp}, we assumed that an abstraction function $f : S \rightarrow \bar{S}$ was given to us. However, since world state is encoded as an image, hand coding $f$ could be difficult. Therefore, instead of assuming $f$ is given directly, we assume that we are given a mechanism for generating training data, $(s,\bar{s}) \in \mathcal{D}$, that can be used to train $f$ as a multi-class classifier. Specifically, we hand-code a function that generates structure instances in simulation given abstract state $\bar{s}$.  Such a generative function is easy to encode for rigid body pick-and-place tasks using object position information from the simulator.

\begin{figure}[h]
  \centering
  \includegraphics[width=0.4\textwidth]{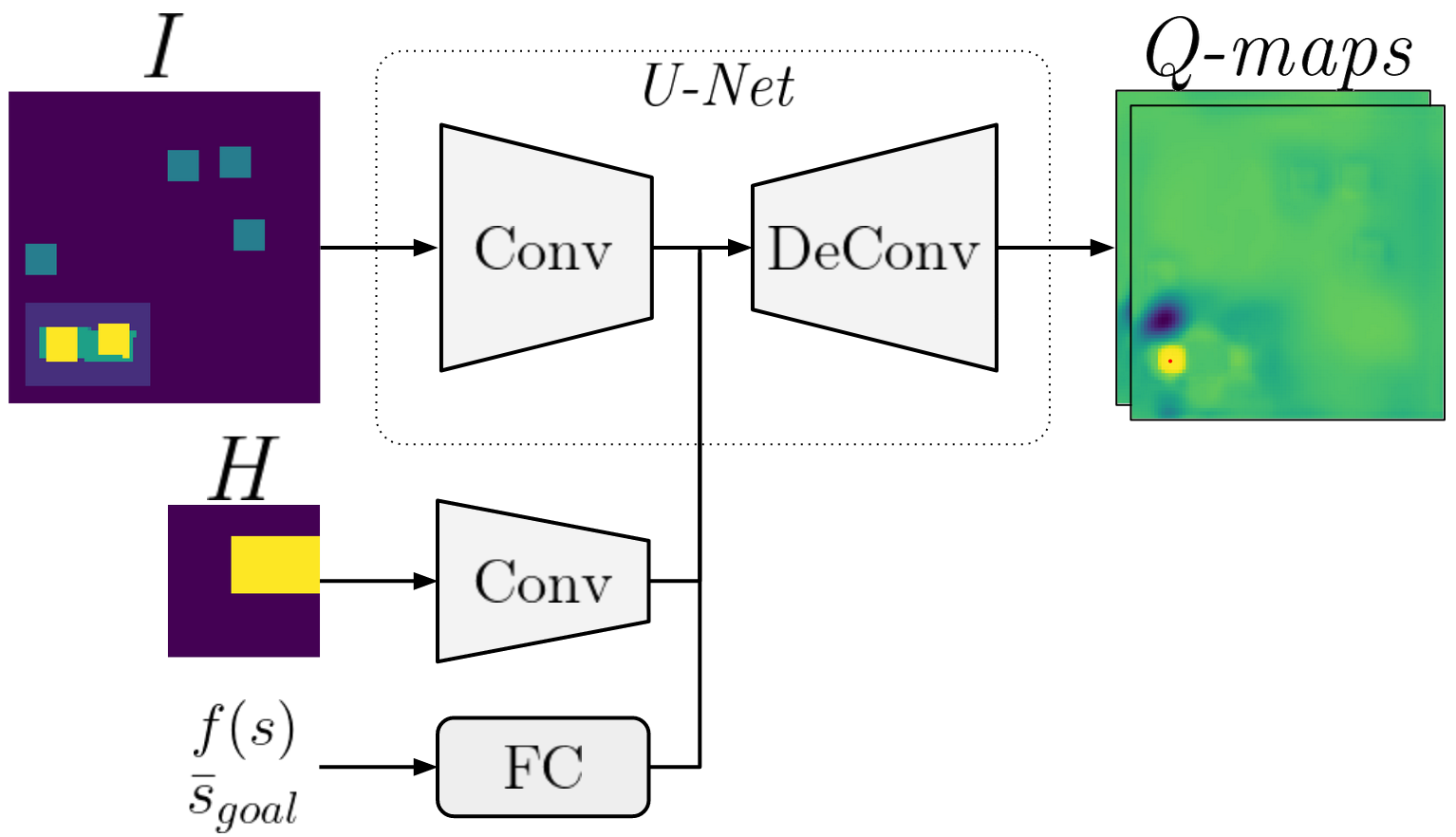}
  \caption{Q-network model has UNet backbone with information on in-hand and abstract state and goal concatenated onto intermediary feature maps. In this example, the robot is holding a long block and selects a place action on the structure in the lower left corner of the workspace.}
  \label{fig:model_architecture}
\end{figure}

\noindent
\underline{$Q$ Network Model Architecture:} Fig.~\ref{fig:model_architecture} shows the model architecture used to learn the goal-conditioned policies. At the upper left, state is input to the network in the form of the world image $I$ and the in-hand image $H$ (Section~\ref{sect:mdp}). At the lower left, we input the abstract goal state $\bar{s}_{goal}$ as well as the current abstract state $f(s)$. The output is a two-channel feature map that encodes the $Q$ value of a spatial action to any pixel position.  The two $Q$-maps correspond to the \textsc{pick} or \textsc{place} actions and is indexed using the boolean flag, $b$. Note that this model takes the current abstract state $f(s)$ as input, which is analogous to the ``achieved goal'' input used in Hindsight Experience Replay~\cite{andrychowicz2017hindsight}. 

\section{EXPERIMENTS}

\subsection{Block Construction Domain}

We explore our approach in a domain where the agent learns to build block structures. We automatically generate the set of subgoal structures shown in Fig.~\ref{fig:height_5_structures} to be those that can be created by combining layers of a single brick, a single roof, or two adjacent cubes. While the size of this set is exponential in height (see Fig.~\ref{fig:lc}a), we only need to program a small set of transition rules once.  The transition rules are created such that an edge exists between two structures if a single block can be added to move between them. We denote any structure that cannot be further built upon, due to a roof block or height limit, as an end-goal structure (i.e. a dead-end node in goal graph). At the start of each episode, the workspace is reset such that all necessary blocks for a given structure are present, and the platform is free of blocks. The block construction domain is simulated in Pybullet \cite{coumans2017pybullet}. 
Abstract states are represented as a one-hot vector over all possible structures plus a boolean flag that indicates that the platform is clear. The state space is $S = S_{world} \times S_{robot}$, where $S_{world} = \mathbb{R}^{1 \times 90 \times 90}$ is an image that describes the state of the world. $S_{robot} = \mathbb{R}^{1 \times 24 \times 24} \times \{0, 1\}$ describes the state of the robot in terms of an ``in-hand'' image that describes the object grasped in the hand and a binary flag that denotes whether the gripper is holding an object. 

An important caveat to note in all the experiments below is that we limit the spatial action space to $x,y$ -- the orientation of the gripper remains fixed throughout. In other words, while we defined $a_{sp} \in SE(2)$ in Section~\ref{sect:mdp}, in these experiments we restrict $a_{sp} \in \mathbb{R}^2$. This limitation made it easier for us to run the experiments below, but we do not view it as a theoretical limitation because the feasibility of policy learning over a $SE(2)$ spatial action space is already well established~\cite{wang2020policy,wu2020spatial}. Given a spatial action at $x,y$, the $z$ dimension of the action is determined using a heuristic based on information from the depth image, as is common in the top-down grasping literature \cite{zeng2020transporter, zeng2018robotic, wang2020policy}.

\subsection{Comparison With Baselines}
\label{sect:comparisons}

\begin{figure*}[t]
    \centering
    \begin{subfigure}[h]{0.24\textwidth}
        \includegraphics[width=\textwidth]{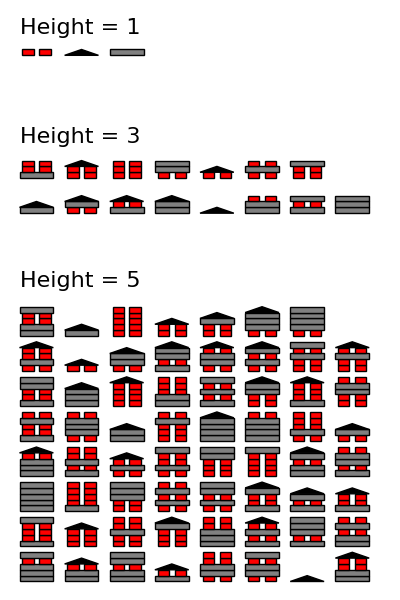}
        \caption{}
    \end{subfigure}
        \begin{subfigure}[h]{0.75\textwidth}
            \includegraphics[width=0.98\textwidth]{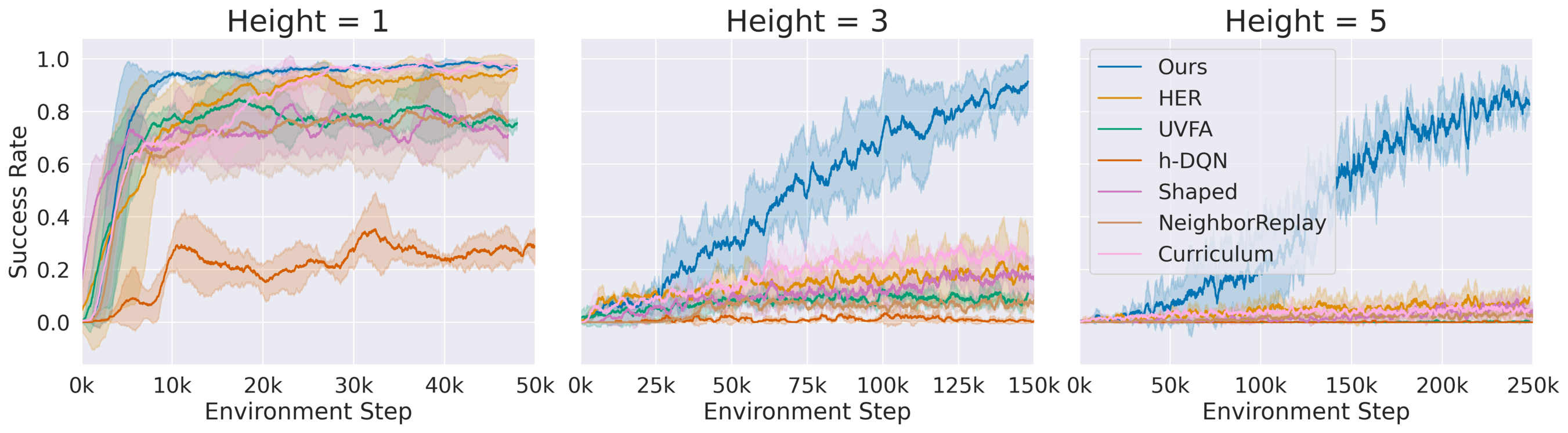}
        \includegraphics[width=0.98\textwidth]{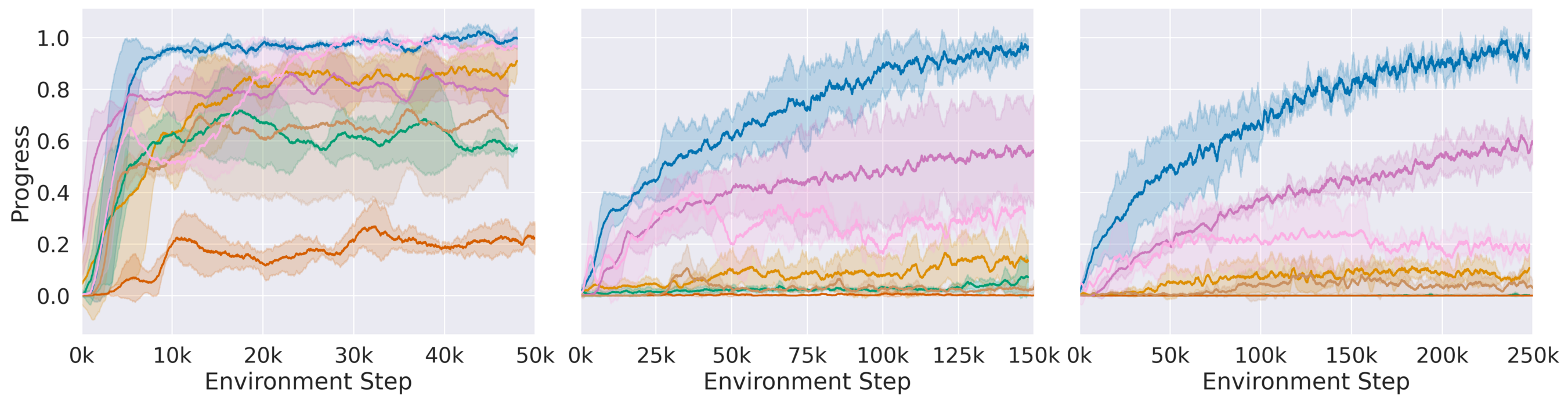}
        \caption{}
    \end{subfigure}
    \vspace{-0.1cm}
    \caption{Comparison of our method to baselines (b) on end goal distributions illustrated in (a). The upper row shows success rate during training, and the lower row shows the average progress (i.e. fraction of blocks in end-goal structure that are placed correctly). Shaded regions denote standard deviation over three runs.}
    \label{fig:lc}
    \vspace{-0.5cm}
\end{figure*}

Here, we evaluate the ability of our approach to learn a set of subgoal policies that can reach any one of a set of end-goal structures. Recall from section~\ref{sect:mdp} that $\Lambda$ is the probability distribution over end-goal abstract states that defines the learning objective. We evaluate our method over three different distributions $\Lambda$ with successively larger support, as shown in Fig.~\ref{fig:lc}a. For each of these three classes, we define $\Lambda_i$ (the probability distribution over end-goals of maximum height $i$) to be the uniform distribution over the structures in that set. We use Q-learning with the version of $\epsilon$-greedy exploration described in section~\ref{sect:policy_learning} where $\epsilon$ varies linearly between $1$ and $0$ during training.

\noindent
\underline{Multi-Goal Baselines:}  We compare our method to three baselines that do not require an structured goal space, Universal Value Function Approximators (UVFA), Hindsight Experience Replay (HER), and h-DQN. In UVFA, we instantiate the same goal conditioned $Q$ network we use for subgoal learning (Fig.~\ref{fig:model_architecture}), but train that network directly without goal relabeling~\cite{schaul2015universal}. End-goals are sampled in exactly the same way they are in Algorithm~\ref{alg:alg1}, and UVFA is conditioned on the sampled end-goal. HER is the same as UVFA except that the method incorporates the standard goal relabeling strategy proposed in~\cite{andrychowicz2017hindsight}. h-DQN is our implementation of~\cite{kulkarni2016hierarchical}, where a top-level network learns to predict appropriate subgoals for the low-level goal-conditioned network. h-DQN selects subgoals from the same set of abstract states as our method, and uses a two-layer MLP to predict useful subgoals for a given end-goal.  These baselines, which are not provided subgoals by goal graph, do see the abstract state at every timestep, which provides useful information. To fully evaluate the effectiveness of our method, we introduce more baselines in the next paragraph that do benefit from the structure of the goal graph.

\noindent
\underline{Structured Learning Baselines:} We compare against three additional baselines, all of which leverage the structure of the abstract goal space in some way.  First, the shaped reward baseline uses the same algorithm as UVFA, except that we provide a shaped reward signal based on the high level plan.  At the start of each episode, an abstract end-goal state, $\bar{s}_{end}$, is sampled, and the path in the goal graph is generated, \{${\bar{s}_1, \bar{s}_2, \dots, \bar{s}_N}$\} where $\bar{s}_N = \bar{s}_{end}$.  At each time step $t$, the agent receives a reward of $+i$ if it is at the $i$-th subgoal in the path (e.g. $\bar{s}_t = \bar{s}_i$), and 0 otherwise. Thus, even though the agent is always conditioned on the end-goal it receives a step-wise reward signal to guide its actions.  The next baseline is called neighbor replay, which is similar to HER, but instead of relabelling achieved abstract states as goal states, it relabels nearby abstract states (e.g. neighbors in the goal graph) as goal states.  The final baseline is a UVFA agent trained with a curriculum, where the end-goal distribution changes overtime from simple to more difficult.  We implement the Sequential curriculum method from \cite{li2020towards}, wherein the agent first learns to stack a single block, then two, and so on.  The curriculum transitions between levels once the agent has achieved over 90\% success rate on the end-goal distribution (e.g. all structures composed of $n$ blocks or less) imposed by the curriculum.

\noindent
\underline{Results and Discussion:} Our method outperforms the baselines, and the performance gap increases with the maximum height of the structures (Fig. \ref{fig:lc}b). The other methods can only reach a fraction of the 63 height-5 structures, whereas our method obtains a 86\% success rate.  In comparison to UVFA and HER, our method performs comparably on height 1 structures (in fact, it is equivalent to UVFA for single block structures).  However, for more complex end-goal distributions, our method greatly outperforms, which we hypothesize is due to the exploration problem. In our method, the graph planner focuses the agent's exploration on nearby goals, increasing the chances of receiving a reward. On the other hand, UVFA and HER are often attempting to reach goals that require many individual block placements.  HER can only relabel achieved states as goals, so we do not expect it to noticeably improve exploration capabilities.  While the meta-controller in h-DQN, which uses a two-level learning hierarchy, could potentially simplify the task assigned to the low-level policy, it is unable to select subgoals that are unknown but reachable within a few time steps.  Knowing the abstract transitions between goals that have not yet been seen is important for having an effective exploration policy for the block construction domain.  The considerable performance boost that our method achieves over the multi-goal baselines demonstrates that the effort of hand-coding goal structure is justified for the block construction domain.

Our method also outperforms the structured learning baselines on challenging end-goal distributions.  On height 5, the success rates of the baselines are low; however, the fraction of blocks correctly placed (lower row of Fig.~\ref{fig:lc}b) are higher in comparison to the multi-goal baselines.  Shaped rewards achieves an average progress of 60\% after 250k steps on height 5 structures, compared to an average progress of over 95\% for our method.  We hypothesize that our method out-competes shaped rewards because learning local policies represents a simpler learning problem.   While curriculum performs comparably to shaped rewards early in training, the performs flat-lines or decreases later.  The set of new goals introduced by the curriculum increases with each level and the agent is unable to overcome this distributional shift to build structures that require more than 2 correct block placements.  The neighbor replay method performed poorly overall.  Even though the relabelling uses nearby abstract states, the chances that the relabelling provides a reward signal is low due to the improbability of accidentally building structures that are nearby in abstract space.  We believe this comparison demonstrates that our method is best at taking advantage of the abstract goal space to solve a challenging multi-goal manipulation problem; our method's implicit curriculum guides exploration, while the locality of the policies improves learning speed and representational capacity.

\subsection{Transfer to Novel Structures }
\label{sect:transfer}
A desirable feature for a multi-goal policy learning method is the ability to transfer knowledge to new tasks.  In our block construction domain, the goal distribution represents only a small fraction of possible structure configurations and object types.  Thus, it would be valuable if our agent could be quickly re-trained to achieve a modified goal distribution. In this section, we evaluate our method's ability to transfer knowledge to goal distributions that include novel objects.

\begin{figure}[H]
    \centering
    \includegraphics[width=0.4\textwidth]{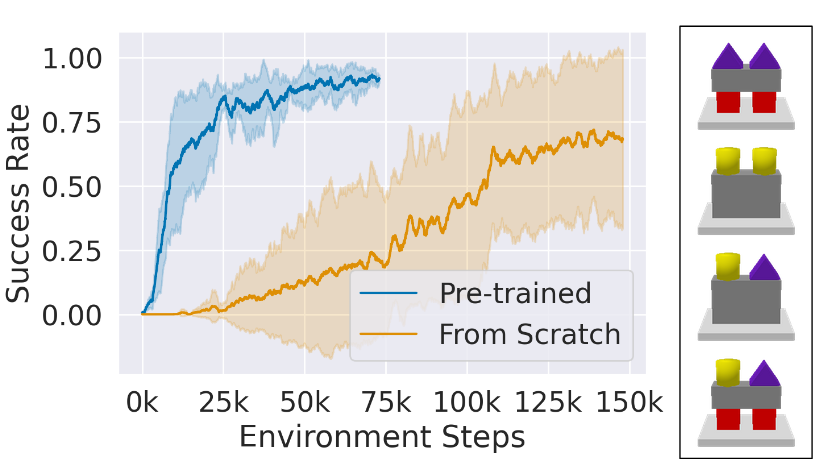}
    \caption{Forward transfer to structures with novel objects. Pre-trained (blue) is our method using a Q-network that was pre-trained on all height 5 structures until convergence, From Scratch (orange) is our method starting with an untrained Q-network. Both methods are trained on the four end-goal structures shown of the right.  The success rate of building these structures is plotted during training, with shading denoting the standard deviation over four independent seeds.}
    \label{fig:transfer}
\end{figure}

\noindent
\underline{Results and Discussion:} Fig.~\ref{fig:transfer} shows learning curves on a novel end-goal distribution.   The new end-goal distribution includes four structures of height 3 that include two novel objects: a small roof (purple) and a cylinder (yellow).  We report the learning curve of our method when the network has been pre-trained to convergence on the height 5 end-goal distribution, compared to when the network is not pre-trained.  When pre-trained our method reached 90\% success rate within 50k environment steps.  Our method is able to transfer to the new goals quickly for two reasons: the structured policy representation means that it can re-use known subgoal policies to build the first two layers; the Q-network has priors on picking and placing that guide exploration even with novel objects.

\subsection{Evaluation in Simulator}
Here, we show the results of training our method to convergence in the simulator.  In Table \ref{tab:sim-breakdown}, performance on all 63 end-goal structures of $\Lambda_5$ is evaluated after training for 500k environment steps.  The average success rate over all these structures is 93\%.  In general, lower success rate is observed on structures that involve several layers of cubes stacked on top of each other.  This is understandable since stacking cubes requires more precision for \textsc{PLACE} actions.  

\begin{table}[H]
    \centering
    \resizebox{0.48\textwidth}{!}{
        \begin{tabular}{cccccccccccc}
	\toprule
	 & 
	\includegraphics[width=0.25in]{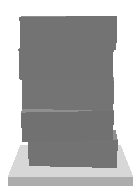} & 
	\includegraphics[width=0.25in]{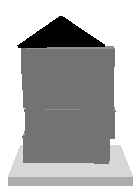} & 
	\includegraphics[width=0.25in]{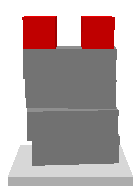} & 
	\includegraphics[width=0.25in]{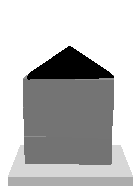} & 
	\includegraphics[width=0.25in]{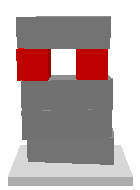} & 
	\includegraphics[width=0.25in]{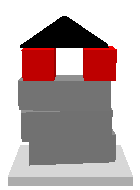} & 
	\includegraphics[width=0.25in]{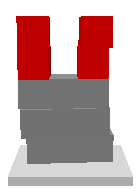} & 
	\includegraphics[width=0.25in]{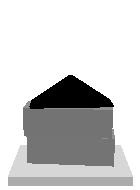} & 
	\includegraphics[width=0.25in]{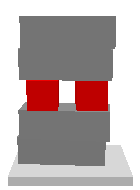} & 
	\includegraphics[width=0.25in]{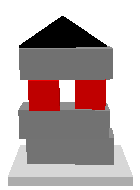} & 
	\includegraphics[width=0.25in]{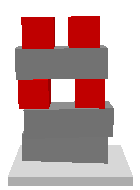} \\
	\midrule 
 
	\textit{Success} & 
	99\% & 	97\% & 	98\% & 	98\% & 	96\% & 	95\% & 	58\% & 	99\% & 	97\% & 	93\% & 	94\% \\ 
	
	\textit{} & 
	$\pm0$\% & 	$\pm 1$\% & 	$\pm 0$\% & 	$\pm 0$\% & 	$\pm 3$\% & 	$\pm 4$\% & 	$\pm 42$\% & 	$\pm 1$\% & 	$\pm 3$\% & 	$\pm 4$\% & 	$\pm 7$\% \\ 


    &&&&&&&&&&& \\ 
	 & 
	\includegraphics[width=0.25in]{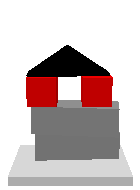} & 
	\includegraphics[width=0.25in]{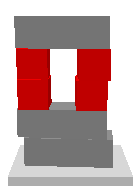} &
	\includegraphics[width=0.25in]{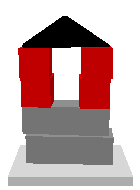} & 
	\includegraphics[width=0.25in]{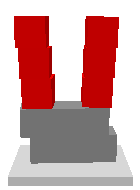} & 
	\includegraphics[width=0.25in]{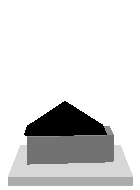} & 
	\includegraphics[width=0.25in]{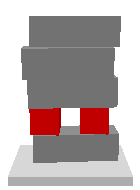} & 
	\includegraphics[width=0.25in]{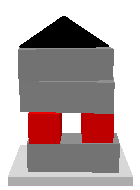} & 
	\includegraphics[width=0.25in]{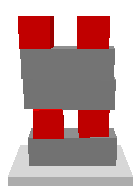} & 
	\includegraphics[width=0.25in]{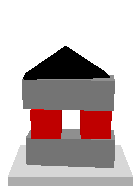} & 
	\includegraphics[width=0.25in]{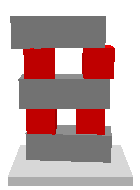} & 
	\includegraphics[width=0.25in]{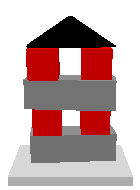} \\
	\midrule 
 
	\textit{Success} & 
	98\% & 	96\%  & 93\% & 	90\% & 	99\% & 	97\% & 	95\% & 	94\% & 	98\% & 	95\% & 	93\% \\ 
	
	\textit{} & 
	$\pm 2$\% & 	$\pm 3$\%  & $\pm 2$\% & $\pm10$\% & 	$\pm 1$\% & 	$\pm 2$\% & 	$\pm 3$\% & 	$\pm 3$\% & 	$\pm 1$\% & 	$\pm 2$\% & 	$\pm 3$\%  \\ 


    &&&&&&&&&&& \\ 
	 & 
	\includegraphics[width=0.25in]{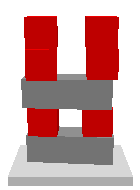} & 
	\includegraphics[width=0.25in]{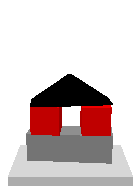} &
	\includegraphics[width=0.25in]{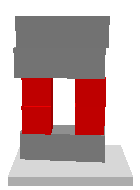} & 
	\includegraphics[width=0.25in]{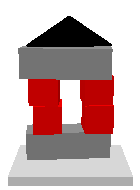} &
	\includegraphics[width=0.25in]{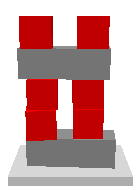} & 
	\includegraphics[width=0.25in]{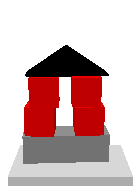} & 
	\includegraphics[width=0.25in]{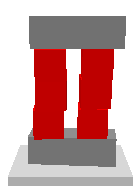} & 
	\includegraphics[width=0.25in]{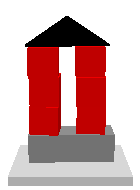} & 
	\includegraphics[width=0.25in]{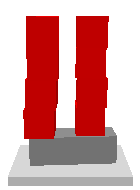} & 
	\includegraphics[width=0.25in]{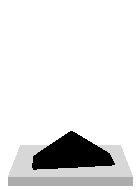} & 
	\includegraphics[width=0.25in]{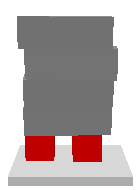} \\
	\midrule 
 
	\textit{Success} & 
	93\% & 	99\% & 	96\% & 	96\% & 90\% & 	97\% & 	93\% & 	93\% & 	93\% & 	100\% & 	92\% \\ 

	\textit{} & 
	$\pm 5$\% & 	$\pm 1$\% & 	$\pm 3$\% & 	$\pm 2$\% & $\pm 9$\% & $\pm 1$\% & 	$\pm 5$\% & 	$\pm 3$\% & 	$\pm 6$\% & 	$\pm 0$\% & 	$\pm 7$\%  \\ 

    &&&&&&&&&&& \\ 
	 & 
	\includegraphics[width=0.25in]{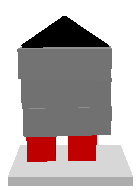} & 
	\includegraphics[width=0.25in]{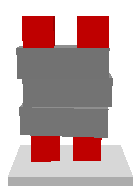} & 
	\includegraphics[width=0.25in]{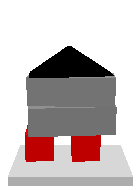} & 
	\includegraphics[width=0.25in]{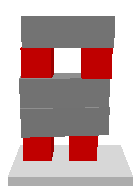} & 
	\includegraphics[width=0.25in]{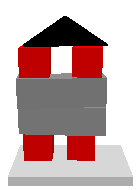} & 
	\includegraphics[width=0.25in]{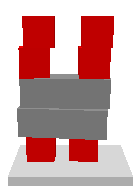} &
	\includegraphics[width=0.25in]{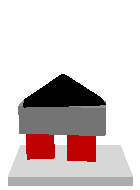} & 
	\includegraphics[width=0.25in]{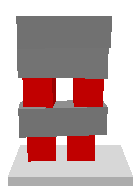} & 
	\includegraphics[width=0.25in]{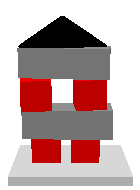} & 
	\includegraphics[width=0.25in]{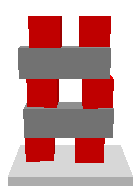} & 
	\includegraphics[width=0.25in]{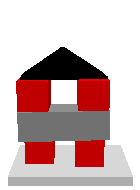} \\ 
	\midrule 
 
	\textit{Success} & 
	92\% & 	95\% & 	97\% & 	89\% & 	91\% & 	92\% & 97\% & 	96\% & 	95\% & 	92\% & 	95\% \\ 

	\textit{} & 
	$\pm 4$\% & 	$\pm 0$\% & 	$\pm 2$\% & 	$\pm 7$\% & 	$\pm 7$\% & 	$\pm 5$\%  & $\pm 3$\% & $\pm 1$\% & 	$\pm 4$\% & 	$\pm 6$\% & 	$\pm 5$\% 	 \\ 

    &&&&&&&&&&& \\ 
	 & 
	\includegraphics[width=0.25in]{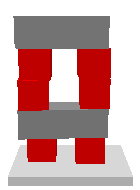} & 
	\includegraphics[width=0.25in]{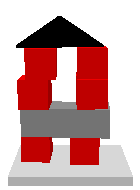} & 
	\includegraphics[width=0.25in]{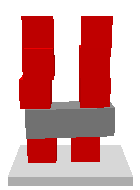} & 
	\includegraphics[width=0.25in]{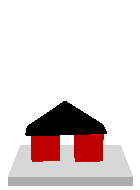} & 
	\includegraphics[width=0.25in]{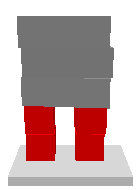} & 
	\includegraphics[width=0.25in]{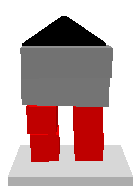} & 
	\includegraphics[width=0.25in]{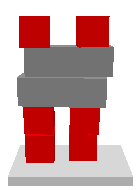} & 
	\includegraphics[width=0.25in]{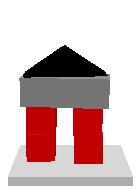} &
	\includegraphics[width=0.25in]{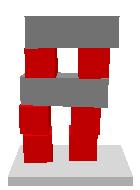} & 
	\includegraphics[width=0.25in]{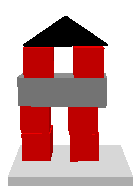} & 
	\includegraphics[width=0.25in]{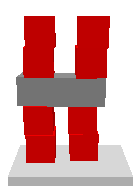} \\
	\midrule 
 
	\textit{Success} & 
	89\% & 	92\% & 	91\% & 	97\% & 	91\% & 	87\% & 	90\% & 	92\% & 89\% & 	85\% & 	86\%  \\ 

	\textit{} & 
	$\pm 10$\% & 	$\pm 6$\% & 	$\pm 4$\% & 	$\pm 2$\% & 	$\pm 7$\% & 	$\pm 7$\% & 	$\pm 7$\% & 	$\pm 4$\% & $\pm 5$\% & $\pm 10$\% & 	$\pm 18$\% \\ 
	
    &&&&&&&&&&& \\ 
	 & 
	\includegraphics[width=0.25in]{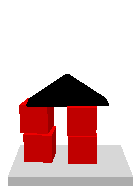} & 
	\includegraphics[width=0.25in]{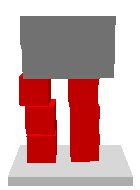} & 
	\includegraphics[width=0.25in]{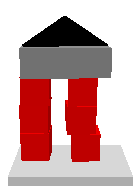} & 
	\includegraphics[width=0.25in]{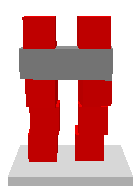} & 
	\includegraphics[width=0.25in]{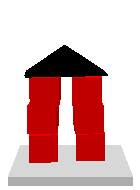} & 
	\includegraphics[width=0.25in]{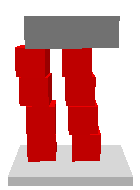} & 
	\includegraphics[width=0.25in]{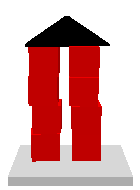} & 
    \includegraphics[width=0.25in]{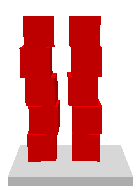} & \\
	\midrule 
    
	\textit{Success} & 
    97\% & 	93\% & 	91\% & 	91\% & 	96\% & 	87\% & 	87\% & 	89\% & & & \\
    
	\textit{} & 
    $\pm 3$\% & 	$\pm 4$\% & 	$\pm 5$\% & 	$\pm 6$\% & 	$\pm 3$\% & 	$\pm 6$\% & 	$\pm 5$\% & 	$\pm 5$\% \\

	\bottomrule
\end{tabular}

    }
    \caption{Performance on 63 end goal structures from $\Lambda_5$, after training for 500k environment steps in simulator.  Success is reported as the percentage of runs that the agent reached the structure.  Results are shown as the average and standard deviation over three independently trained agents. For each agent, success rates are evaluated by performing 100 trajectories for each end goal structure.}
    \label{tab:sim-breakdown}
\end{table}

\subsection{Evaluation on Real-World Robot}

\begin{figure}[h]
    \centering
     \hfill
    \begin{subfigure}[b]{0.26\textwidth}
         \centering
         \includegraphics[width=\textwidth]{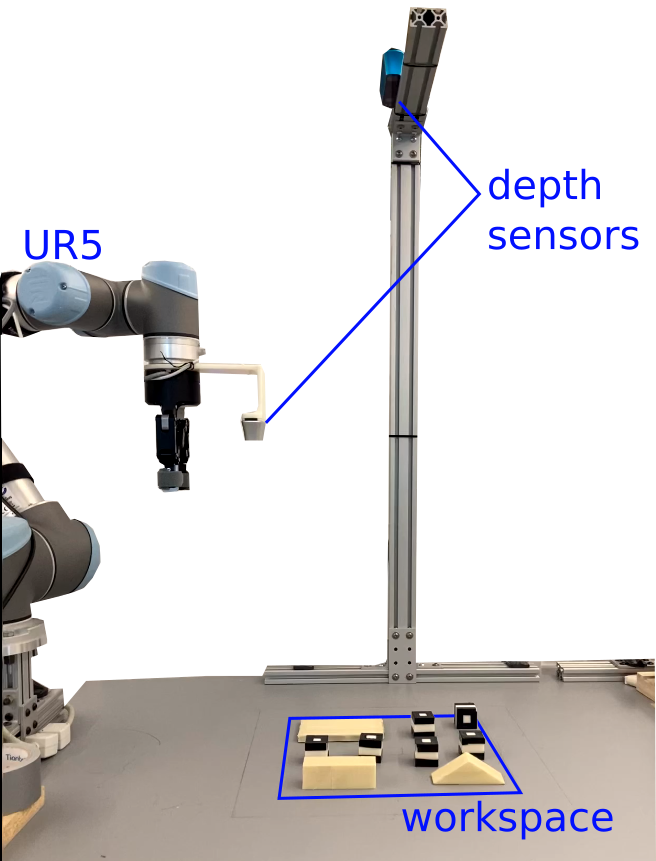}
         \caption{}
     \end{subfigure}
     \hfill
     \begin{subfigure}[b]{0.14\textwidth}
         \centering
         \includegraphics[width=\textwidth]{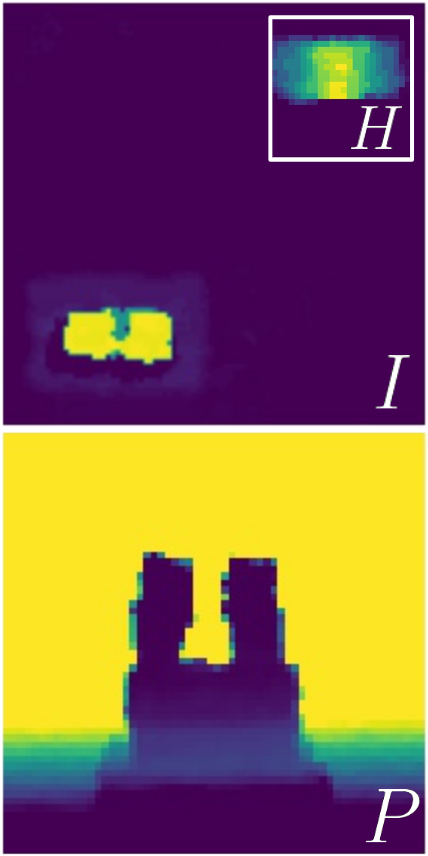}
         \caption{}
     \end{subfigure}
     \hfill
    \vspace{-0.1cm}
    \caption{(a) setup for experiments on real-world robot after reset, (b) observations from the real-world system as the agent is about to place a roof to complete the structure. Images, $H$ and $I$, are captured with the top-mounted depth sensor, while the depth image of the platform, $P$, is captured by wrist-mounted depth sensor after moving the arm accordingly.}
    \label{fig:real-robot-setup}
\end{figure}

\begin{figure}[h]
    \centering
    \includegraphics[width=\linewidth]{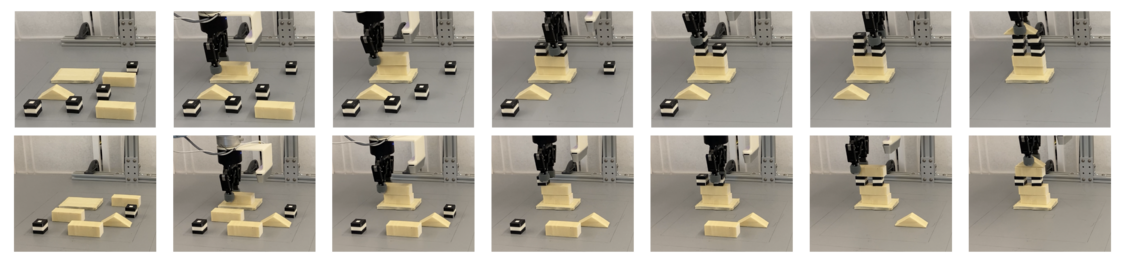}
    \caption{Two examples of block construction performed on our UR5 robotic system, trajectories start on leftmost column.}
    \label{fig:real-trajectories}
    \vspace{-0.5cm}
\end{figure}

\noindent
\underline{Setup:} We performed experiments on a UR5 robot equipped with a Robotiq two-finger gripper which was mounted on a flat table top. Depth sensor measurements were provided by two Occipital Structure sensors -- one mounted above the table pointing down and the other mounted at the wrist (setup is shown in Fig.~\ref{fig:real-robot-setup}a). We trained the agent for 500k environment steps in the simulator before evaluating on the real robotic system. Fig.~\ref{fig:real-trajectories} shows two examples of our agent building structures with the real robot. 

\begin{table*}[t]
    \centering
    \resizebox{0.95\textwidth}{!}{%
    \begin{tabular}{cllllllllll}
     \toprule
      &
     \includegraphics[width=0.4in]{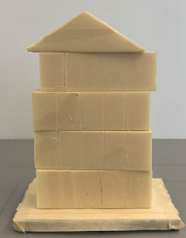} &  
     \includegraphics[width=0.4in]{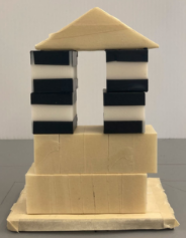} &  
     \includegraphics[width=0.4in]{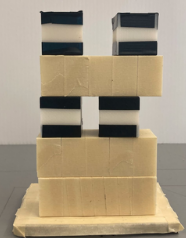} &  
     \includegraphics[width=0.4in]{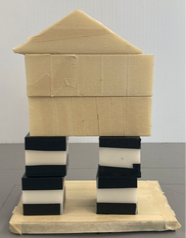} &  
     \includegraphics[width=0.4in]{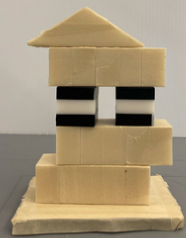} &  
     \includegraphics[width=0.4in]{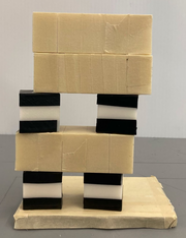} &  
     \includegraphics[width=0.4in]{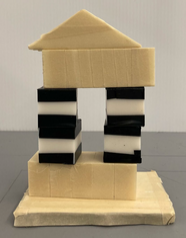} &  
     \includegraphics[width=0.4in]{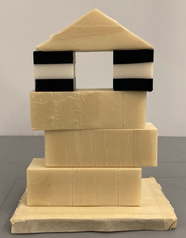} &  
     \includegraphics[width=0.4in]{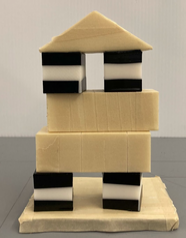} &  
     \includegraphics[width=0.4in]{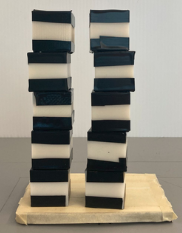} \\
     \midrule
     
    \textit{Success} &
     90\%  &
     100\% &
     80\%  &
     90\%  &
     100\% &
     80\%  &
     90\%  &
     80\%  &
     100\% &
     50\%  \\

     \textit{Progress} &
     98\%  $\pm 6$\% &
     100\% $\pm 0$\% &
     90\%  $\pm23$\% &
     96\%  $\pm13$\% &
     100\% $\pm 0$\% &
     91\%  $\pm23$\% &
     96\%  $\pm14$\% &
     90\%  $\pm21$\% &
     100\% $\pm 0$\% &
     83\%  $\pm25$\%\\
     \bottomrule
\end{tabular}

    }
    \caption{We ran the policy learned by our method 10 times for each of the ten different end-goal structures shown above. The row labeled \textit{Success} shows the percentage of runs in which the structure was successfully built. The row labeled \textit{Progress} is the percentage of blocks correctly placed in each structure, and the standard deviation is reported.}
    \label{tab:real-robot-performance}
    \vspace{-0.5cm}
\end{table*}

\noindent
\underline{Sim-to-real transfer:} To deploy our learned policy on the real world robot, two modifications were made to our method.  First, data augmentation was used during training to bridge the gap between depth images in the simulator and depth images in the real world.  We found the best results by augmenting the simulator's depth images with Perlin noise (magnitude 0.05) followed by Gaussian blurring ($\sigma$ sampled between 0.5 to 1). Additionally, to account for the fact that real life blocks were not perfectly axis-aligned, we added a small amount of orientation noise to the block placement upon environment resets.

The second, and more significant, modification to our method is the introduction of a learned abstraction function.  Recall that our method relies on an abstraction function to determine the current abstract state, and decide when a policy has reached its subgoal.  In the simulator, the abstraction function leverages privileged access to object poses.  For the real world, we implemented a classifier network which takes depth images of the structures from the side (image $P$ from Fig.~\ref{fig:real-robot-setup}b) and outputs the corresponding abstract state.  The network was trained in a supervised manner with cross-entropy loss using 150k pairs of structure images and abstract state labels collected using the fully trained agent in the simulator.  Thus, when running on the real robot, side view images of the structure were taken every time step to produce an abstract state encoding via the trained classifier.

\noindent
\underline{Results and Discussion:} Success rates from our real robot runs are shown in Table~\ref{tab:real-robot-performance}. We performed 10 runs on each of the 10 different structures shown in the table (100 attempts at block construction in total). We report the percentage of runs that each structure is built successfully (Success) and the percentage of blocks in the structure that were correctly placed (Progress). The results suggest that our method can be very successful on a real robotic system. It is important to emphasize that these structures were created using a single goal-conditioned policy -- we did not learn a separate policy for each structure.  Most failures resulted from slightly inaccurate block placements which caused structures to fall. This is especially evident with the ``two towers'' structure on the far right where there was only a 50\% success rate, with an average 8.3 correct block placements out of 10.


\section{CONCLUSIONS}
\label{sec:conclusions}

\noindent
\underline{Discussion:} This paper demonstrates that a simple approach to hierarchical policy learning involving hand-coding high level structure can enable large scale multi-goal policy learning. As far as we know, there is no other method that can learn a single policy that can construct more than a hundred different block structures using images as state. One concern is that this method could only be relevant to block construction tasks and not to more practical manipulation applications. However, in related work, Wang et al.~\cite{wang2022equivariant} show that policy learning in a spatial action space can indeed be an effective way to solve practical robotics manipulation applications and we believe that this work would could be extended similarly. 

\noindent
\underline{Limitations:} An important failure mode we observed occurs when the agent is attempting to construct the two towers shown on the right side of Table~\ref{tab:real-robot-performance} but the towers have collapsed because the agent did not align successive blocks sufficiently accurately on top of one another. The problem here is that our method does not back up reward through multiple subgoal policies. Since policy learning only reasons about the stability of the next successive block, it fails to recognize that a misplaced block early in the structure could cause failure that only occurs much later during construction. Fundamentally, this results from a defect in how abstract states are defined -- had they been more restrictive in terms of block alignment, this failure would not occur. However, anticipating all possible failure modes of a hand-coded abstraction is infeasible for complex domains; a possible avenue for handling this problem is to enable the abstraction to adapt online as a result of end-to-end feedback. 





\newpage
\bibliographystyle{IEEEtran}
\bibliography{IEEEabrv, references}

\begin{thebibliography}{10}
\providecommand{\url}[1]{#1}
\csname url@rmstyle\endcsname
\providecommand{\newblock}{\relax}
\providecommand{\bibinfo}[2]{#2}
\providecommand\BIBentrySTDinterwordspacing{\spaceskip=0pt\relax}
\providecommand\BIBentryALTinterwordstretchfactor{4}
\providecommand\BIBentryALTinterwordspacing{\spaceskip=\fontdimen2\font plus
\BIBentryALTinterwordstretchfactor\fontdimen3\font minus
  \fontdimen4\font\relax}
\providecommand\BIBforeignlanguage[2]{{%
\expandafter\ifx\csname l@#1\endcsname\relax
\typeout{** WARNING: IEEEtran.bst: No hyphenation pattern has been}%
\typeout{** loaded for the language `#1'. Using the pattern for}%
\typeout{** the default language instead.}%
\else
\language=\csname l@#1\endcsname
\fi
#2}}

\bibitem{plappert2018multi}
M.~Plappert, M.~Andrychowicz, A.~Ray, B.~McGrew, B.~Baker, G.~Powell,
  J.~Schneider, J.~Tobin, M.~Chociej, P.~Welinder, \emph{et~al.}, ``Multi-goal
  reinforcement learning: Challenging robotics environments and request for
  research,'' \emph{arXiv preprint arXiv:1802.09464}, 2018.

\bibitem{kaelbling2010hierarchical}
L.~P. Kaelbling and T.~Lozano-P{\'e}rez, ``Hierarchical planning in the now,''
  in \emph{Workshops at the Twenty-Fourth AAAI Conference on Artificial
  Intelligence}, 2010.

\bibitem{srivastava2014combined}
S.~Srivastava, E.~Fang, L.~Riano, R.~Chitnis, S.~Russell, and P.~Abbeel,
  ``Combined task and motion planning through an extensible planner-independent
  interface layer,'' in \emph{2014 IEEE international conference on robotics
  and automation (ICRA)}.\hskip 1em plus 0.5em minus 0.4em\relax IEEE, 2014,
  pp. 639--646.

\bibitem{pong2018temporal}
V.~Pong, S.~Gu, M.~Dalal, and S.~Levine, ``Temporal difference models:
  Model-free deep rl for model-based control,'' \emph{arXiv preprint
  arXiv:1802.09081}, 2018.

\bibitem{andrychowicz2017hindsight}
M.~Andrychowicz, F.~Wolski, A.~Ray, J.~Schneider, R.~Fong, P.~Welinder,
  B.~McGrew, J.~Tobin, P.~Abbeel, and W.~Zaremba, ``Hindsight experience
  replay,'' \emph{arXiv preprint arXiv:1707.01495}, 2017.

\bibitem{zhao18energy}
R.~Zhao and V.~Tresp, ``Energy-based hindsight experience prioritization,'' in
  \emph{2nd Annual Conference on Robot Learning, CoRL 2018, Z{\"{u}}rich,
  Switzerland, 29-31 October 2018, Proceedings}, 2018, pp. 113--122.

\bibitem{zhao19entropy}
R.~Zhao, X.~Sun, and V.~Tresp, ``Maximum entropy-regularized multi-goal
  reinforcement learning,'' in \emph{Proceedings of the 36th International
  Conference on Machine Learning, {ICML} 2019, 9-15 June 2019, Long Beach,
  California, {USA}}, 2019, pp. 7553--7562.

\bibitem{li2020towards}
R.~Li, A.~Jabri, T.~Darrell, and P.~Agrawal, ``Towards practical multi-object
  manipulation using relational reinforcement learning,'' in \emph{2020 IEEE
  International Conference on Robotics and Automation (ICRA)}.\hskip 1em plus
  0.5em minus 0.4em\relax IEEE, 2020, pp. 4051--4058.

\bibitem{funk2022learn2assemble}
N.~Funk, G.~Chalvatzaki, B.~Belousov, and J.~Peters, ``Learn2assemble with
  structured representations and search for robotic architectural
  construction,'' in \emph{Conference on Robot Learning}.\hskip 1em plus 0.5em
  minus 0.4em\relax PMLR, 2022, pp. 1401--1411.

\bibitem{lin2022efficient}
Y.~Lin, A.~S. Wang, E.~Undersander, and A.~Rai, ``Efficient and interpretable
  robot manipulation with graph neural networks,'' \emph{IEEE Robotics and
  Automation Letters}, vol.~7, no.~2, pp. 2740--2747, 2022.

\bibitem{nair18visual}
A.~Nair, V.~Pong, M.~Dalal, S.~Bahl, S.~Lin, and S.~Levine, ``Visual
  reinforcement learning with imagined goals,'' in \emph{Advances in Neural
  Information Processing Systems 31: Annual Conference on Neural Information
  Processing Systems 2018, NeurIPS 2018, December 3-8, 2018, Montr{\'{e}}al,
  Canada}, 2018, pp. 9209--9220.

\bibitem{nasiriany2019planning}
S.~Nasiriany, V.~H. Pong, S.~Lin, and S.~Levine, ``Planning with
  goal-conditioned policies,'' \emph{arXiv preprint arXiv:1911.08453}, 2019.

\bibitem{janner2018reasoning}
M.~Janner, S.~Levine, W.~T. Freeman, J.~B. Tenenbaum, C.~Finn, and J.~Wu,
  ``Reasoning about physical interactions with object-oriented prediction and
  planning,'' \emph{arXiv preprint arXiv:1812.10972}, 2018.

\bibitem{florensa18autmatic}
C.~Florensa, D.~Held, X.~Geng, and P.~Abbeel, ``Automatic goal generation for
  reinforcement learning agents,'' in \emph{Proceedings of the 35th
  International Conference on Machine Learning, {ICML} 2018,
  Stockholmsm{\"{a}}ssan, Stockholm, Sweden, July 10-15, 2018}, 2018, pp.
  1514--1523.

\bibitem{zhang2020automatic}
Y.~Zhang, P.~Abbeel, and L.~Pinto, ``Automatic curriculum learning through
  value disagreement,'' \emph{Advances in Neural Information Processing
  Systems}, vol.~33, 2020.

\bibitem{florensa17reverse}
C.~Florensa, D.~Held, M.~Wulfmeier, M.~Zhang, and P.~Abbeel, ``Reverse
  curriculum generation for reinforcement learning,'' in \emph{1st Annual
  Conference on Robot Learning, CoRL 2017, Mountain View, California, USA,
  November 13-15, 2017, Proceedings}, 2017, pp. 482--495.

\bibitem{sukhbaatar18intrinsic}
S.~Sukhbaatar, Z.~Lin, I.~Kostrikov, G.~Synnaeve, A.~Szlam, and R.~Fergus,
  ``Intrinsic motivation and automatic curricula via asymmetric self-play,'' in
  \emph{6th International Conference on Learning Representations, {ICLR} 2018,
  Vancouver, BC, Canada, April 30 - May 3, 2018, Conference Track Proceedings},
  2018.

\bibitem{matiisen20teacher}
T.~Matiisen, A.~Oliver, T.~Cohen, and J.~Schulman, ``Teacher-student curriculum
  learning,'' \emph{{IEEE} Trans. Neural Networks Learn. Syst.}, vol.~31,
  no.~9, pp. 3732--3740, 2020.

\bibitem{mukherjee2021reactive}
S.~Mukherjee, C.~Paxton, A.~Mousavian, A.~Fishman, M.~Likhachev, and D.~Fox,
  ``Reactive long horizon task execution via visual skill and precondition
  models,'' in \emph{2021 IEEE/RSJ International Conference on Intelligent
  Robots and Systems (IROS)}.\hskip 1em plus 0.5em minus 0.4em\relax IEEE,
  2021, pp. 5717--5724.

\bibitem{kaelbling17learning}
L.~P. Kaelbling and T.~Lozano{-}P{\'{e}}rez, ``Learning composable models of
  parameterized skills,'' in \emph{2017 {IEEE} International Conference on
  Robotics and Automation, {ICRA} 2017, Singapore, Singapore, May 29 - June 3,
  2017}.\hskip 1em plus 0.5em minus 0.4em\relax {IEEE}, 2017, pp. 886--893.

\bibitem{james20learning}
S.~D. James, B.~Rosman, and G.~Konidaris, ``Learning portable representations
  for high-level planning,'' in \emph{Proceedings of the 37th International
  Conference on Machine Learning, {ICML} 2020, 13-18 July 2020, Virtual Event},
  ser. Proceedings of Machine Learning Research, vol. 119.\hskip 1em plus 0.5em
  minus 0.4em\relax {PMLR}, 2020, pp. 4682--4691.

\bibitem{stolle2002learning}
M.~Stolle and D.~Precup, ``Learning options in reinforcement learning,'' in
  \emph{International Symposium on abstraction, reformulation, and
  approximation}.\hskip 1em plus 0.5em minus 0.4em\relax Springer, 2002, pp.
  212--223.

\bibitem{bacon2017option}
P.-L. Bacon, J.~Harb, and D.~Precup, ``The option-critic architecture,'' in
  \emph{Proceedings of the AAAI Conference on Artificial Intelligence},
  vol.~31, no.~1, 2017.

\bibitem{vezhnevets2017feudal}
A.~S. Vezhnevets, S.~Osindero, T.~Schaul, N.~Heess, M.~Jaderberg, D.~Silver,
  and K.~Kavukcuoglu, ``Feudal networks for hierarchical reinforcement
  learning,'' in \emph{International Conference on Machine Learning}.\hskip 1em
  plus 0.5em minus 0.4em\relax PMLR, 2017, pp. 3540--3549.

\bibitem{konidaris2018skills}
G.~Konidaris, L.~P. Kaelbling, and T.~Lozano-Perez, ``From skills to symbols:
  Learning symbolic representations for abstract high-level planning,''
  \emph{Journal of Artificial Intelligence Research}, vol.~61, pp. 215--289,
  2018.

\bibitem{watter15embed}
M.~Watter, J.~T. Springenberg, J.~Boedecker, and M.~A. Riedmiller, ``Embed to
  control: {A} locally linear latent dynamics model for control from raw
  images,'' in \emph{Advances in Neural Information Processing Systems 28:
  Annual Conference on Neural Information Processing Systems 2015, December
  7-12, 2015, Montreal, Quebec, Canada}, C.~Cortes, N.~D. Lawrence, D.~D. Lee,
  M.~Sugiyama, and R.~Garnett, Eds., 2015, pp. 2746--2754.

\bibitem{levine18learning}
S.~Levine, P.~Pastor, A.~Krizhevsky, J.~Ibarz, and D.~Quillen, ``Learning
  hand-eye coordination for robotic grasping with deep learning and large-scale
  data collection,'' \emph{Int. J. Robotics Res.}, vol.~37, no. 4-5, pp.
  421--436, 2018.

\bibitem{mnih2015human}
V.~Mnih, K.~Kavukcuoglu, D.~Silver, A.~A. Rusu, J.~Veness, M.~G. Bellemare,
  A.~Graves, M.~Riedmiller, A.~K. Fidjeland, G.~Ostrovski, \emph{et~al.},
  ``Human-level control through deep reinforcement learning,'' \emph{nature},
  vol. 518, no. 7540, pp. 529--533, 2015.

\bibitem{van2016deep}
H.~Van~Hasselt, A.~Guez, and D.~Silver, ``Deep reinforcement learning with
  double q-learning,'' in \emph{Proceedings of the AAAI Conference on
  Artificial Intelligence}, vol.~30, no.~1, 2016.

\bibitem{coumans2017pybullet}
E.~Coumans and Y.~Bai, ``Pybullet, a python module for physics simulation in
  robotics, games and machine learning,'' 2017.

\bibitem{wang2020policy}
D.~Wang, C.~Kohler, and R.~Platt, ``Policy learning in se (3) action spaces,''
  \emph{arXiv preprint arXiv:2010.02798}, 2020.

\bibitem{wu2020spatial}
J.~Wu, X.~Sun, A.~Zeng, S.~Song, J.~Lee, S.~Rusinkiewicz, and T.~Funkhouser,
  ``Spatial action maps for mobile manipulation,'' \emph{arXiv preprint
  arXiv:2004.09141}, 2020.

\bibitem{zeng2020transporter}
A.~Zeng, P.~Florence, J.~Tompson, S.~Welker, J.~Chien, M.~Attarian,
  T.~Armstrong, I.~Krasin, D.~Duong, V.~Sindhwani, \emph{et~al.}, ``Transporter
  networks: Rearranging the visual world for robotic manipulation,''
  \emph{arXiv preprint arXiv:2010.14406}, 2020.

\bibitem{zeng2018robotic}
A.~Zeng, S.~Song, K.-T. Yu, E.~Donlon, F.~R. Hogan, M.~Bauza, D.~Ma, O.~Taylor,
  M.~Liu, E.~Romo, \emph{et~al.}, ``Robotic pick-and-place of novel objects in
  clutter with multi-affordance grasping and cross-domain image matching,'' in
  \emph{2018 IEEE international conference on robotics and automation
  (ICRA)}.\hskip 1em plus 0.5em minus 0.4em\relax IEEE, 2018.

\bibitem{schaul2015universal}
T.~Schaul, D.~Horgan, K.~Gregor, and D.~Silver, ``Universal value function
  approximators,'' in \emph{International conference on machine
  learning}.\hskip 1em plus 0.5em minus 0.4em\relax PMLR, 2015, pp. 1312--1320.

\bibitem{kulkarni2016hierarchical}
T.~D. Kulkarni, K.~Narasimhan, A.~Saeedi, and J.~Tenenbaum, ``Hierarchical deep
  reinforcement learning: Integrating temporal abstraction and intrinsic
  motivation,'' \emph{Advances in neural information processing systems},
  vol.~29, pp. 3675--3683, 2016.

\bibitem{wang2022equivariant}
D.~Wang, R.~Walters, X.~Zhu, and R.~Platt, ``Equivariant $ q $ learning in
  spatial action spaces,'' in \emph{Conference on Robot Learning}.\hskip 1em
  plus 0.5em minus 0.4em\relax PMLR, 2022, pp. 1713--1723.

\end{thebibliography}

\end{document}